\documentclass[11pt]{article}

\usepackage[preprint]{acl}

\usepackage{times}
\usepackage{latexsym}

\usepackage[T1]{fontenc}

\usepackage[utf8]{inputenc}

\usepackage{microtype}

\usepackage{inconsolata}

\title{From Flat Logs to Causal Graphs: Hierarchical Failure Attribution \\ for LLM-based Multi-Agent Systems}

\author{
 \textbf{Yawen Wang\textsuperscript{1,2,3,4}\thanks{These authors contributed equally to this work.}},
 \textbf{Wenjie Wu\textsuperscript{5}}\footnotemark[1],
 \textbf{Junjie Wang\textsuperscript{1,2,3,4}}\thanks{Corresponding authors.},
 \textbf{Qing Wang\textsuperscript{1,2,3,4}}\footnotemark[2],
\\
\\
\textsuperscript{1}Institute of Software, Chinese Academy of Sciences, Beijing, China
\\
\textsuperscript{2}State Key Laboratory of Complex System Modeling and Simulation Technology, Beijing, China
\\
\textsuperscript{3}Science \& Technology on Integrated Information System Laboratory, Beijing, China
\\
\textsuperscript{4}University of Chinese Academy of Sciences, Beijing, China
\\
\textsuperscript{5}Wuhan University of Technology, Wuhan, China
\\
\{yawen2018, junjie, wq\}@iscas.ac.cn,
louis\_wu@whut.edu.cn
}

\usepackage{booktabs}       
\usepackage{amsfonts}       
\usepackage[table]{xcolor}
\usepackage{amsmath}
\usepackage{graphicx}
\usepackage{multirow}
\usepackage{subcaption} 
\usepackage{enumitem}
\usepackage{pifont}
\usepackage{array}     

\newcommand{\tool}{CHIEF}
\newcommand{\website}{https://anonymous.4open.science/r/CHIEF-86B8}

\newcommand{\NA}{--}    

\usepackage{caption}
\usepackage{cleveref}
\usepackage{listings}
\lstdefinestyle{promptstyle}{
  basicstyle=\ttfamily\small,        
  columns=fullflexible,         
  breaklines=true,                
  breakatwhitespace=true, 
  keepspaces=true,                  
  showstringspaces=false,         
  frame=tb,                        
  rulecolor=\color{black!80},                       
  aboveskip=0em,
  belowskip=0.3em,
  xleftmargin=1.5em,
  xrightmargin=1.5em,
  backgroundcolor=\color{gray!5}
}

\begin{document}
\maketitle

\begin{abstract}

LLM-powered Multi-Agent Systems (MAS) have demonstrated remarkable capabilities in complex domains but suffer from inherent fragility and opaque failure mechanisms. 
Existing failure attribution methods, whether relying on direct prompting, costly replays, or supervised fine-tuning, typically treat execution logs as flat sequences. This linear perspective fails to disentangle the intricate causal links inherent to MAS, leading to weak observability and ambiguous responsibility boundaries.
To address these challenges, we propose {\tool}, a novel framework that transforms chaotic trajectories into a structured \emph{hierarchical causal graph}.
It then employs \emph{hierarchical oracle-guided backtracking} to efficiently prune the search space via sybthesized virtual oracles.
Finally, it implements \emph{counterfactual attribution} via a progressive causal screening strategy to rigorously distinguish true root causes from propagated symptoms.
Experiments on \textit{Who\&When} benchmark show that {\tool} outperforms eight strong and state-of-the-art baselines on both agent- and step-level accuracy.
Ablation studies further confirm the critical role of each proposed module.

\end{abstract}

\section{Introduction}
\label{sec:intro}

The advent of Large Language Models (LLMs) has empowered agents with exceptional proficiency in perception~\cite{conf/iclr/ZhengLFL24}, planning~\cite{conf/icml/ErdoganL0MFAKG25}, and reasoning~\cite{journals/corr/abs-2408-07199}. 
Building on these capabilities, Multi-Agent Systems (MASs) have emerged to orchestrate specialized agents, achieving superior performance in complex domains such as software engineering~\cite{journals/pacmse/MaCCZCLLLHL25, conf/iclr/0001LSXTZPSLSTL25} and general-purpose realistic tasks~\cite{conf/iclr/MialonF0LS24, conf/emnlp/YoranAMBPB24}.
However, the integration of autonomous agents alongside diverse tools introduces inherent fragility, with recent studies revealing failure rates up to 86.7\%~\cite{journals/corr/abs-2503-13657}, where errors propagate through intricate dependencies, rendering systems unreliable and opaque.

\begin{figure}[t]
    \centering
    \includegraphics[width=\columnwidth]{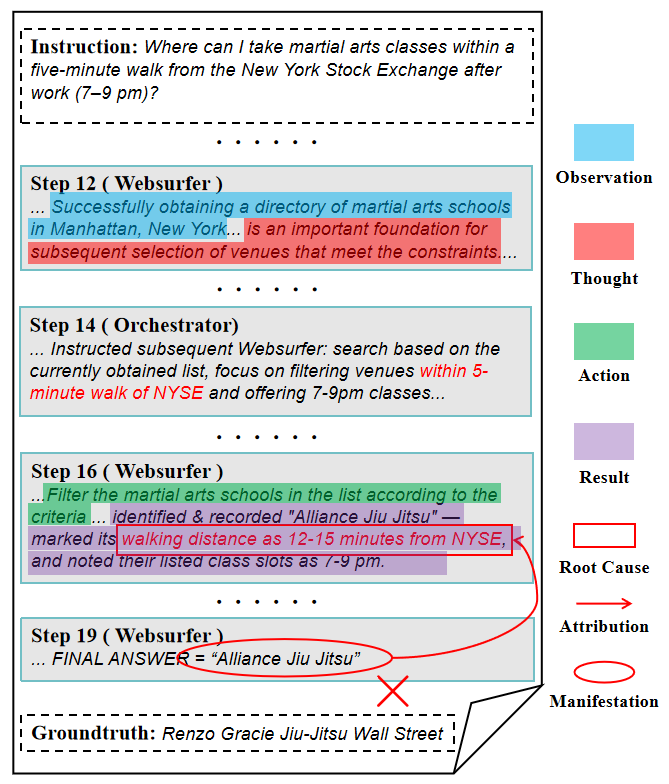}
    \caption{A failure log induced by an unsatisfied constraint.
    Although the \textit{Orchestrator} marking constraints (red, Step 14), the \textit{Websurfer} overlooks these constraints during execution (Step 16).
    }
    \label{fig:log}
    \vspace{-15pt} 
\end{figure}

Consequently, \textit{Failure Attribution}, identifying the root cause and the responsible agent for a task failure, has emerged as a critical yet challenging task. Since the introduction of the \textit{Who\&When} benchmark~\cite{conf/icml/ZhangY0LHZL0W0W25}, various approaches have been proposed to automate this diagnosis.
Direct LLM-based methods~\cite{conf/icml/ZhangY0LHZL0W0W25,journals/corr/abs-2510-04886} often struggle to capture fine-grained causal clues within lengthy contexts; Spectrum-based methods~\cite{journals/corr/abs-2509-13782} incur prohibitive token costs through repeated trajectory replays; Fine-tuning-based methods~\cite{journals/corr/abs-2509-03312,journals/corr/abs-2510-10581} introduce significant training overheads and generalization risks.

More critically, current approaches are limited by treating MAS trajectories as flat sequences, ignoring the inherent structural complexity illustrated in Fig.~\ref{fig:log}.
Unlike simple linear texts, these logs represent a dense intertwining of agents' observations, thoughts, actions, and results, interconnected by rigorous data dependencies (e.g., web search results) and agent interactions (e.g., orchestrator-executor interactions).
This lack of structural abstraction and causal disentanglement leaves existing methods struggling to address three diagnostic obstacles inherent to failure attribution:

\noindent\textbf{Opaque Causal Flows:}
As illustrated in Fig.~\ref{fig:log}, raw MAS logs exhibit a dense entanglement of tool executions, environmental feedback, and inter-agent communications.
Without structural parsing, the implicit dependencies and causal links are submerged in this verbose text, leading to weak observability for attribution analysis.

\noindent\textbf{Sparse Intermediate Supervision:} Unlike traditional software debugging with modular unit tests,
MAS trajectories lack intermediate ground truth. As correctness is observable only at the final outcome, pinpointing the precise failure step within long-horizon trajectories becomes a ``needle in a haystack'' search.

\noindent\textbf{Ambiguous Responsibility Boundaries:} 
As exemplified in Fig.~\ref{fig:log}, a constraint violation (e.g., \textit{``5-minute walk''}) triggers failure (Step 19), yet it is unclear whether the error stems from the orchestrator's omission (Step 14) or the executor's negligence (Step 16). 
This discrepancy between where an error manifests and where it was introduced makes distinguishing root causes from propagated symptoms challenging without causal abstraction.

To address these obstacles, we propose {\tool}, a \textbf{C}ausal \textbf{HIE}rarchical \textbf{F}ailure attribution framework for LLM-based MASs.
Departing from paradigms that treat logs as flat sequences, we explicitly reconstruct chaotic trajectories into a structured graph, transforming attribution into a transparent and systematic divide-and-conquer process.
First,
we construct a \textit{Hierarchical Causal Graph}. {\tool} decomposes tasks into subtasks, applies \textit{Observation-Thought-Action-Result} (\textit{OTAR}) parsing to disentangle intertwined agent behaviors, and models the data dependencies between steps explicitly.
Second,
we propose \textit{Hierarchical Oracle-Guided Backtracking}. This module acts as an organized divide-and-conquer strategy, replacing linear scans with a top-down search.
By verifying subtasks via synthesized virtual oracles, we bypass granular action inspection, efficiently pruning the search space to pinpoint the precise failure step.
Finally,
we implement \textit{Counterfactual Attribution} via a progressive causal screening strategy. By filtering responsibility through causal scope and dependency nature, and applying a deviation-aware check for reversibility, we rigorously distinguish true root causes from propagated symptoms.

Our experimental evaluation leverages the \textit{Who\&When} benchmark~\cite{conf/icml/ZhangY0LHZL0W0W25}, spanning 127 diverse MAS architectures. 
{\tool} demonstrates superior performance, outperforming 8 baselines categorized into four paradigms. 
It delivers 77.59\% agent-level and 29.31\% step-level accuracy on the hand-crafted subset, alongside 76.80\% and 52.00\% on the algorithm-generated subset. 
Furthermore, ablation studies isolate the impact of our proposed modules, verifying that the hierarchical causal graph and virtual oracle are critical for effective failure attribution.
Our contributions are summarized as follows:

\begin{itemize}[noitemsep]
    \item A novel method, {\tool}, which transforms chaotic trajectories into a structured graph, enabling a transparent and systematic divide-and-conquer attribution process.
    \item A \textit{Hierarchical Causal Graph} construction method, establishing a structured foundations that facilitates precise failure attribution and other understanding and analysis for MASs.
    \item Extensive experiments on \textit{Who\&When} benchmark, demonstrating superior performance over existing baselines.

    \item The replication package on our website\footnote{\url{\website}} to support reproducibility and future research.

\end{itemize}

\section{Related Work}
\label{sec:related_work}

\subsection{LLM-based Multi-Agent Systems}
LLM-based MASs achieve superior performance on complex tasks by orchestrating specialized agents~\cite{conf/nips/LiHIKG23, conf/iclr/HongZCZCWZWYLZR24, journals/corr/abs-2308-08155}. Existing frameworks generally fall into two categories: hand-crafted systems (e.g., AutoGen~\cite{journals/corr/abs-2308-08155}, MetaGPT~\cite{conf/iclr/HongZCZCWZWYLZR24}, ChatDev~\cite{conf/acl/QianLLCDL0CSCXL24}) that rely on pre-defined standard operating procedures, and automated systems (e.g., AgentPrune~\cite{journals/corr/abs-2508-05988}, AFlow~\cite{conf/iclr/ZhangXYTCCZCHWZ25}) that autonomously optimize agent roles and topologies.

Despite successes in reasoning and assistant tasks~\cite{conf/icml/ZhugeWKFKS24, conf/iclr/MialonF0LS24}, the intricate orchestration of agents and tools introduces inherent fragility, where errors propagate through opaque dependencies. 
Shifting focus from system construction, we tackle the critical downstream task of automated failure attribution.

\subsection{Failure Attribution for LLM Agents}
Failure attribution is pivotal for system debugging, yet existing methods face distinct limitations.
Early works~\cite{conf/icml/ZhangY0LHZL0W0W25} employ the \textit{LLM-as-a-Judge} paradigm but fail to handle lengthy logs, often yielding $<10\%$ accuracy. While \textit{ECHO}~\cite{journals/corr/abs-2510-04886} introduces hierarchical context and consensus voting, it treats hierarchy merely as static representation, often mistaking visible symptoms for hidden root causes.
Spectrum-based \textit{FAMAS}~\cite{journals/corr/abs-2509-13782} relies on repeated replays for statistical attribution. While effective for long trajectories, it incurs high costs and yields correlations without causal explanations.
Approaches like \textit{AgenTracer}~\cite{journals/corr/abs-2509-03312} and \textit{GraphTracer}~\cite{journals/corr/abs-2510-10581} fine-tune specialized models (e.g., 8B) on synthetic failure data. Despite achieving high accuracy, they demand substantial upfront costs for data generation and model training, posing generalization risks when applied to unseen agent logs.

Distinct from prior works, {\tool} reconstructs a hierarchical causal graph that enables efficient one-pass reasoning, yielding promising attribution without costly replays or additional training.

\section{Problem Formulation}


\noindent\textbf{LLM MAS.} 
We follow the standard turn-based LLM-based MAS protocol~\cite{conf/iclr/HongZCZCWZWYLZR24, journals/corr/abs-2308-08155}, where exactly one agent performs an action at each timestep.
The MAS is defined as $\mathcal{M} = \langle \mathcal{N}, \mathcal{S}, \mathcal{A}, \mathcal{P} \rangle$, consisting of $N$ agents $\mathcal{N}=\{ i \}_{i=0}^N$. 
At step $t$, active agent $i_t$ executes an action $a_t$,
transitioning the system from $s_t$ to $s_{t+1}$ via the transition probability $\mathcal{P}(s_{t+1} | s_t, a_t)$. 
This yields a trajectory $\tau = \{ s_t, (i_t, a_t) \}_{t=0}^T$, with outcome $Z(\tau) \in \{0, 1\}$ ($1$ denotes failure).

\noindent\textbf{Failure Attribution.} 
For a failed trajectory $\tau$ ($Z(\tau)=1$), the problem of \textit{Failure Attribution} is to identify the specific agent-step pair $(i, t)$ as the root cause of the failure.
We first define the \textit{decisive error indicator} $\Delta_{i,t}(\tau) = \mathbb{I}( Z(\tilde{\tau}^{(i,t)}) = 0 )$, where $\tilde{\tau}^{(i,t)}$ is the counterfactual trajectory with agent $i$'s action at step $t$ corrected. 
$\Delta_{i,t}=1$ implies that the correction leads to success ($Z(\tilde{\tau}^{(i,t)}) = 0$), identifying $(i,t)$ as a decisive error, and 0 otherwise.

Practically, a trajectory may contain multiple decisive errors due to error propagation.
The root cause is the earliest decisive error in the temporal sequence~\cite{conf/icml/ZhangY0LHZL0W0W25}:
\vspace{-8pt}
\begin{equation}
    (i^*, t^*) = \operatorname*{arg\,min}_{(i,t) : \Delta_{i,t}(\tau) = 1} t
\end{equation}
Consequently, the objective is to construct a mapping $\mathcal{F}$ that accurately recovers the ground truth root cause $(i^*, t^*)$ from the raw trajectory $\tau$.

\section{Method}
\label{sec:method}

\begin{figure*}[t]
    \centering
    \includegraphics[width=\textwidth]{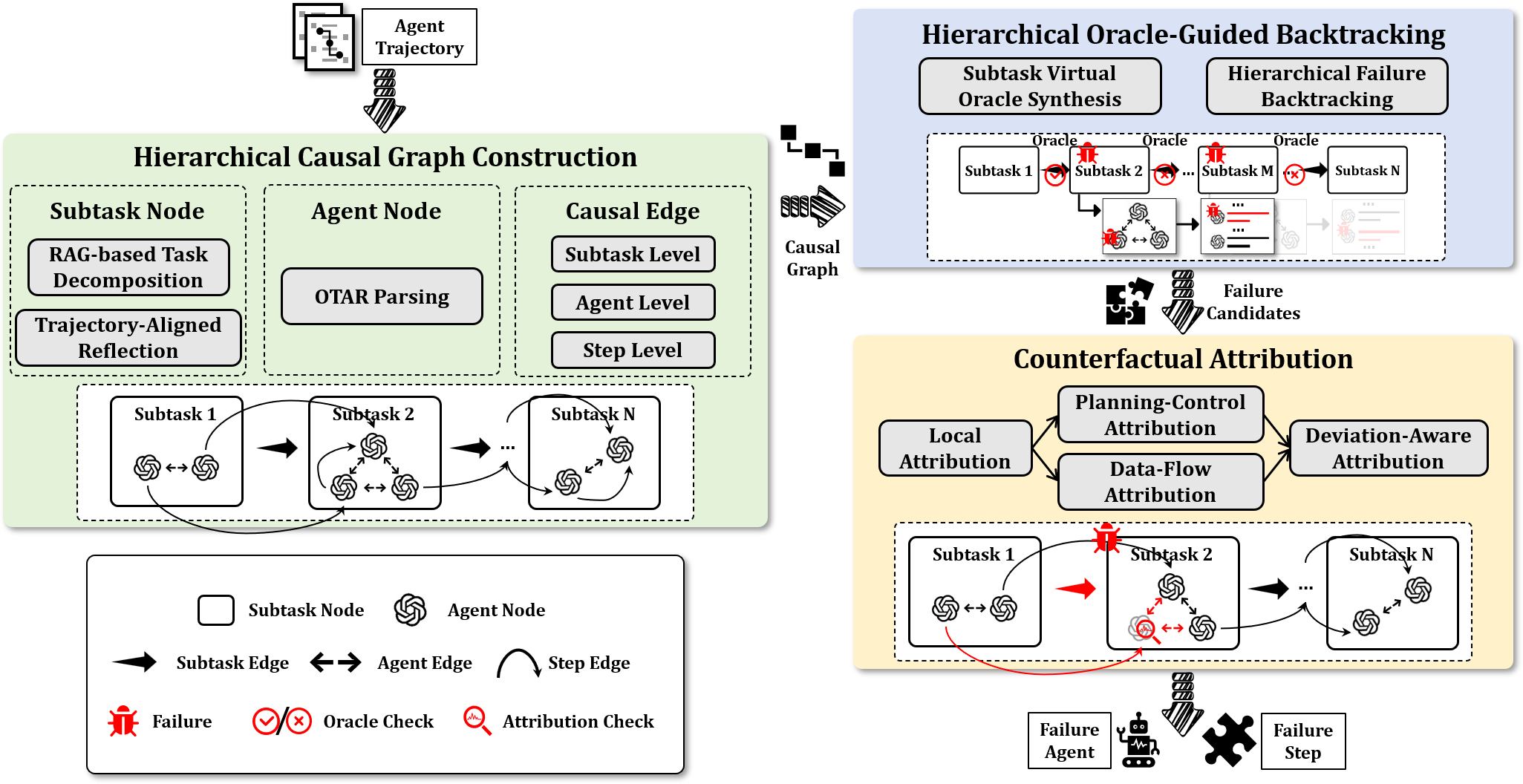}
    \caption{
    The overview of {\tool}.
    }
    \label{fig:overview}
    \vspace{-15pt}
\end{figure*}

To resolve the complexity of MAS diagnosis, we propose a causal hierarchical failure attribution framework. As shown in Fig.~\ref{fig:overview}, we transform the attribution into a structured and divide-and-conquer reasoning process via three phases: (1) \textit{Hierarchical Causal Graph Construction} to parse flat trajectories; (2) \textit{Hierarchical Oracle-Guided Backtracking} to identify error candidates top-down; and (3) \textit{Counterfactual Attribution} to distinguish root causes from propagated symptoms.

\subsection{Hierarchical Causal Graph Construction}

Raw MAS logs, while appearing as flat text sequences, inherently have a latent execution topology.
It interweaves environmental feedback, high-level planning, and low-level execution, etc., all interconnected by implicit data/control dependencies, agent interactions and logical/temporal relations.
Since errors propagate precisely along these hidden pathways rather than linear text, we construct a \textit{Hierarchical Causal Graph (HCG)}, denoted as $\mathcal{G} = (\mathcal{V}, \mathcal{E})$, to explicitly mirror this structure.

\subsubsection{Hierarchical Node}
Nodes $\mathcal{V} = \mathcal{V}_{sub} \cup \mathcal{V}_{agt}$ abstract the trajectory into semantic units at two granularities:


\textbf{Subtask Node} 
$S_k \in \mathcal{V}_{sub}$ is defined as a high-level logical abstraction representing a distinct task phase, characterized by the structured attributes detailed in Fig.~\ref{subfig:graph_subtask}. To partition the raw trajectory $\tau$ into a sequence $\{S_k\}_{k=0}^K$ that is both logically sound and faithful to the execution log, we employ a two-step process:


\noindent(1) RAG-based Task Decomposition. To ensure rationality, we construct a knowledge base derived from existing benchmarks (i.e., GAIA~\cite{conf/iclr/MialonF0LS24}, AssistantBench~\cite{conf/emnlp/YoranAMBPB24}), 
which are populated with human-annotated step-by-step solution exemplars. We leverage Retrieval-Augmented Generation (RAG) to retrieve relevant decomposition prototypes, using them as few-shot prompts to guide the LLM in generating a modular task plan.
Implementation details and example prompts are provided in Appx.~\ref{appx:task_decomposition}.

\noindent(2) Trajectory-Aligned Reflection. To prevent hallucinated task decomposition, a reflection mechanism 
verifies the alignment between the generated $S_k$ and the raw log $\tau$. Mismatched subtasks undergo iterative refinement to ensure the decomposition strictly follows the actual execution flow.

\textbf{Agent Node} 
$a \in \mathcal{V}_{agt}$ is defined as an atomic execution unit representing a specific agent instance within a subtask.
To rigorously characterize the agent's behavior, we assign structured attributes to each node using the \textit{OTAR} tuple: $\langle \text{Observation}, \text{Thought}, \text{Action}, \text{Result} \rangle$, extended from the \textit{TAR} schema~\cite{journals/corr/abs-2506-18824}. 
The components of these attributes are extracted from the raw trajectory $\tau$ via an LLM-based parser. 
We detail \textit{OTAR} parsing in Appx.~\ref{appx:prompt_otar_parsing} and illustrate agent node attributes in Fig.~\ref{subfig:graph_agent}.


\subsubsection{Causal Edge Construction}
The edge set $\mathcal{E} = \mathcal{E}_{sub} \cup \mathcal{E}_{agt} \cup \mathcal{E}_{step}$ captures causal dependencies across three hierarchical levels. 
Specifically, \textit{Subtask Edges ($\mathcal{E}_{sub}$)} link adjacent subtasks to model high-level logical progression of the task, while \textit{Agent Edges ($\mathcal{E}_{agt}$)} connect agent nodes to represent inter-agent collaboration. 
At the finest granularity, we explicitly construct \textit{Step Edges ($\mathcal{E}_{step}$)} by mapping data flow dependencies (e.g., variable references) between execution steps. This multi-level connectivity establishes the necessary pathways for tracing error propagation.


We construct different edges based on the edge type.
Subtask and agent edges utilize \textit{Counterfactual Patterns} $\Phi$, i.e., $\text{Bias}(u) \xrightarrow{\Phi} \text{Anomaly}(v)$, linking latent upstream deviations $\text{Bias}(u)$ to observable downstream errors $\text{Anomaly}(v)$. 
In contrast, step edges serve as data snapshots, explicitly recording the exact upstream outputs and downstream inputs. 
By combining them, we provide comprehensive evidence for the subsequent backtracking phase. 
We detail the implementation in Appx.~\ref{appx:edge_construction} and illustrate edge attributes in Fig.~\ref{subfig:graph_edge}

\subsection{Hierarchical Oracle-Guided Backtracking}
To pinpoint the root cause $(a^*, x^*)$ (agent-step pair) within the graph $\mathcal{G}$, we propose a two-phase strategy comprising \textit{Subtask Virtual Oracle Synthesis} and \textit{Hierarchical Failure Backtracking}. 
We first synthesize a virtual oracle $\mathcal{O}_k$ for each subtask node $S_k$, serving as an intermediate supervision to verify execution correctness at subtask level. We then execute a top-down backtracking process to progressively narrow down the failure scope from coarse-grained subtask to fine-grained agent steps.

\subsubsection{Subtask Virtual Oracle Synthesis}
For each subtask node $S_k \in \mathcal{V}_{sub}$, we synthesize a virtual oracle $\mathcal{O}_k$ serving as the ideal intermediate verifier.
We formalize it as a structured semantic tuple: $\mathcal{O}_k = \langle \mathcal{G}_{sub}, \mathcal{P}_{pre}, \mathcal{E}_{key}, \mathcal{C}_{acc} \rangle$.
$\mathcal{G}_{sub}$ defines subtask's specific \textit{Goal} derived from the task instruction, while $\mathcal{P}_{pre}$ outlines the \textit{Preconditions} dependent 
on upstream outputs and environmental states. Furthermore, $\mathcal{E}_{key}$ highlights the \textit{Key Evidence} that must be verified during reasoning, and $\mathcal{C}_{acc}$ establishes the \textit{Acceptance Criteria} for determining subtask's success.

We model the synthesis of $\mathcal{O}_k$ as a generation function $\mathcal{F}_{gen}$ parameterized by an LLM:
\vspace{-8pt}
\begin{equation}
    \mathcal{O}_k = \mathcal{F}_{gen}(\mathcal{I}, \mathcal{O}_{<k}, \tau_{>k}, \tau_{s})
\end{equation}
The inputs comprise the task instruction $\mathcal{I}$, the history of preceding oracles $\mathcal{O}_{<k}$, the subsequent unprocessed trajectory $\tau_{>k}$ (i.e., the remaining steps following the completed subtasks), and similar task decomposition examples $\tau_{s}$ retrieved via RAG.
The implementation details are presented in Appx.~\ref{appx:oracle_synthesis}.

\subsubsection{Hierarchical Failure Backtracking}

Based on the causal graph and the synthesized oracles, we execute a top-down backtracking procedure to progressively narrow the failure scope. This process identifies failure candidates at three levels:

\noindent\textbf{Subtask Level.} 
To trace the failure source, we traverse subtask nodes in reverse topological order. 
For each subtask $S_k$, we employ an LLM-based semantic evaluator $\mathcal{F}_{eval}$ to get binary decisions, by comparing its actual execution output against the oracle's \textit{Goal} ($\mathcal{G}_{sub}$) and \textit{Acceptance Criteria} ($\mathcal{C}_{acc}$).
An output of $1$ signifies discrepancies, identifying the subtask as a failure candidate:
\vspace{-8pt}
\begin{equation}
    \mathcal{C}_{sub} = \{ S_k \in \mathcal{V}_{sub} \mid \mathcal{F}_{eval}(S_k, \langle \mathcal{G}_{sub}, \mathcal{C}_{acc} \rangle) = 1 \}
\end{equation}


\noindent\textbf{Agent Level.} 
For each candidate subtask $S_k \in \mathcal{C}_{sub}$, we drill down to its constituent agents.  
We apply the semantic evaluator $\mathcal{F}_{eval}$ to assess the consistency of each agent's ($a \in S_k$) \textit{OTAR} tuple with the oracle's \textit{Preconditions} ($\mathcal{P}_{pre}$) and \textit{Key Evidence} ($\mathcal{E}_{key}$). 
Agents exhibiting biases (i.e., $\mathcal{F}_{eval}$ outputting $1$) are identified as candidates:
\vspace{-8pt}
\begin{equation}
    \mathcal{C}_{agt} = \{ a \in S_k \mid \mathcal{F}_{eval}(a_{otar}, \langle \mathcal{P}_{pre}, \mathcal{E}_{key} \rangle) = 1 \}
\end{equation}

\noindent\textbf{Step Level.} 
Given the candidate agent $a \in \mathcal{C}_{agt}$ within subtask $S_k \in \mathcal{C}_{sub}$, we perform a final scrutiny on its executed steps $\mathcal{X}$. 
We employ $\mathcal{F}_{eval}$ to verify each step $x \in \mathcal{X}$, cross-referencing its actual execution details (e.g., input/output) against the agent's summarized \textit{OTAR} ($a_{otar}$) and the subtask's oracle constraints ($\mathcal{O}_k$). 
Steps exhibiting deviations (i.e., $\mathcal{F}_{eval}$ outputting $1$) are pinpointed as failure candidates:
\vspace{-8pt}
\begin{equation}
    \mathcal{C}_{step} = \{ x \in \mathcal{X} \mid \mathcal{F}_{eval}(x, \langle a_{otar}, \mathcal{O}_k \rangle) = 1 \}
\end{equation}
The detailed implementation of LLM-based semantic evaluator $\mathcal{F}_{eval}$ is presented in Appx.~\ref{appx:hierarchical_backtracking}.

\subsection{Counterfactual Attribution}
\label{sec:attribution}
Having identified failure candidates, this phase attributes them to their true origins. 
To systematically disentangle error propagation paths,
we employ a progressive causal screening strategy. 
This approach distinguishes responsibility based on causal scope (local vs. non-local) via \textit{Local Attribution}, and dependency nature (control vs. data) via \textit{Planning-Control} and \textit{Data-Flow Attribution}.
Finally, \textit{Deviation-Aware Attribution} serves as a validity filter, pruning transient errors that are subsequently self-corrected.
The rationale for each stage follows, while technical implementation details are provided in Appx.~\ref{appx:attribution}.

\subsubsection{Local Attribution}
\label{sec:local}

This stage verifies whether the localized error originates at step $x$ itself, rather than being propagated from upstream.
We identify the upstream causal triggers $\mathcal{S}_{cause}$ by applying the counterfactual patterns $\Phi$ bound to the dependency edges:
\vspace{-8pt}
\begin{equation}
    \mathcal{S}_{cause} = \{ x' \in \text{Pre}(x) \mid \text{Bias}(x') \xrightarrow{\Phi} \text{Anomaly}(x) \}
\end{equation}
$\mathcal{S}_{cause}$ represents any prior steps $x' \in \text{Pre}(x)$ that causally account for the current error at $x$.
We attribute the error based on this set:
If $\mathcal{S}_{cause} = \emptyset$, it implies that step $x$ received valid inputs yet produced an incorrect output. Thus, we attribute the root cause locally to $x$.
If $\mathcal{S}_{cause} \neq \emptyset$, the error is propagated from upstream. We exclude $x$ and proceed to the \textit{Planning-Control} and \textit{Data-Flow} stages to pinpoint the non-local source.

\subsubsection{Planning-Control Attribution}
\label{sec:planning_control}
If the error is determined to be non-local, we first investigate control flow errors, which frequently manifest as redundant cyclic behaviors. 
Since these cycles often obscure the boundary between planning and execution errors, our core idea is to distinguish whether the agent attempts to adapt its plan or simply fails to execute a valid plan.
To decouple responsibility, we aggregate the repeating sequence of steps into a \textit{Loop Group} and check the specific rationale behind the iterative re-planning and re-execution steps.
We assign planner's (orchestrator) responsibility when the planner generates semantically identical thoughts or commands despite receiving repeated error signals, indicating failing to update the control flow.
Conversely, we identify executor's responsibility when the planner actively proposes valid strategy shifts (e.g., modifying tool arguments or APIs) to break the loop, yet the executor consistently yields abnormal results.
Fig.~\ref{fig:loop} provides concrete examples for both failure modes.

\subsubsection{Data-Flow Attribution}
Beyond control flow, non-local errors can also propagate through data dependencies.
We reconstruct the error propagation path using step-level data flow edges $\mathcal{E}_{step}$ and explicit variable references in \textit{OTAR} tuples.
We audit data consistency along this path, backtracking to pinpoint the specific step where valid upstream inputs are first corrupted into an abnormal result.
This effectively disentangles the data generator (the true root cause, e.g., a hallucinated fact) from downstream data propagators (symptoms, e.g., calculation errors based on bad inputs), ensuring attribution targets the origin rather than the manifestation.

\subsubsection{Deviation-Aware Attribution}
MASs often exhibit self-correcting capabilities.
To strictly distinguish root causes from transient fluctuations, we assess the \textit{Causal Reversibility} of identified errors. 
For an upstream suspect step $x_t$, we examine the subsequent trajectory $\tau_{>t}$ for evidence of self-correction. If a later step $x_{t+k}$ successfully re-satisfies the oracle criteria (i.e., the system returns to a valid state), the initial deviation at $x_t$ is deemed reversible and assigned no responsibility.
Consequently, we prioritize the attribution to irreversible errors.

\section{Experiment Setup}
\label{sec:experiment}

\subsection{Benchmark}

Our evaluation utilizes \textit{Who\&When}~\cite{conf/icml/ZhangY0LHZL0W0W25}, to our knowledge, currently the sole public benchmark for MAS failure attribution.
Derived from general-purpose tasks in GAIA~\cite{conf/iclr/MialonF0LS24} and AssistantBench~\cite{conf/emnlp/YoranAMBPB24}, the dataset comprises 184 failure logs divided into two subsets:
(1) an algorithm-generated subset containing 126 logs from 126 diverse architectures created by CaptainAgent~\cite{journals/corr/abs-2405-19425}; 
and (2) a hand-crafted subset containing 58 logs from the Magnetic-One~\cite{journals/corr/abs-2411-04468} system.
Ground truth reliability is ensured through multi-round consensus annotation by human experts.

\subsection{Baselines}
We compare {\tool} against eight representative approaches categorized into four paradigms:

\noindent\textit{\textbf{Random}}: A lower-bound baseline that randomly selects an agent and a step from the log.

\noindent\textit{LLM-based Prompting}: We include three strategies proposed in the \textit{Who\&When} benchmark~\cite{conf/icml/ZhangY0LHZL0W0W25}: (1) \textit{\textbf{All-at-once}} (direct prediction), (2) \textit{\textbf{Step-by-step}} (sequential judgment), and (3) \textit{\textbf{Binary-search}} (recursive narrowing). 
We additionally include \textit{\textbf{ECHO}}~\cite{journals/corr/abs-2510-04886}, which combines hierarchical context extraction with consensus voting.
Notably, \textit{ECHO} utilizes hierarchy primarily for static context representation and objective consistency checks, rather than construct a causal graph for top-down backtracking.

\noindent\textit{Spectrum-based Method}: We include \textit{\textbf{FAMAS}}~\cite{journals/corr/abs-2509-13782}, which works by statistically analyzing variations across repeated trajectory replays.

\noindent\textit{Fine-tuning-based Methods}: We include two 8B-parameter fine-tuned models: \textit{\textbf{AgenTracer}}~\cite{journals/corr/abs-2509-03312}, trained via multi-granular reinforcement learning on counterfactual replays; and \textit{\textbf{GraphTracer}}~\cite{journals/corr/abs-2510-10581}, which utilizes information dependency graphs to construct synthetic samples for supervised attribution.

\begin{table*}[t]
\centering
\small 
\renewcommand{\arraystretch}{1}
\setlength{\tabcolsep}{8pt} 

\caption{
Main results on the \textit{Who\&When} benchmark. 
We report agent- and step-level accuracy (\%) across both subsets. 
Each cell reports two values: the left corresponds to the setting \textbf{w/ $\mathcal{G}$}, and the right corresponds to \textbf{w/o $\mathcal{G}$}, where $\mathcal{G}$ indicates access to the task ground truth (i.e., correct outcome).
Best results are in \textbf{bold}.}
\label{tab:exp-results}

\begin{tabular}{llcccc} 
\toprule
\multirow{2}{*}{\textbf{Type}} & \multirow{2}{*}{\textbf{Method}} & \multicolumn{2}{c}{\textbf{Hand-Crafted Dataset}} & \multicolumn{2}{c}{\textbf{Algorithm-Generated Dataset}} \\ 
\cmidrule(lr){3-4} \cmidrule(lr){5-6}
 &  & \textit{Agent-level} $\uparrow$ & \textit{Step-level} $\uparrow$ & \textit{Agent-level} $\uparrow$ & \textit{Step-level} $\uparrow$ \\ 
\midrule

\multirow{1}{*}{Heuristic} 
 & Random         & 12.00 / 12.00 & 4.20 / 4.20   & 29.10 / 29.10 & 19.10 / 19.10 \\
\midrule

\multirow{4}{*}{\shortstack[l]{LLM-based\\Prompting}}
 & All-at-Once    & 50.00 / 48.28 & 5.17 / 5.17   & 61.11 / 59.52 & 13.49 / 15.87 \\
 & Step-by-Step   & 36.00 / 34.30 & 6.60 / 6.90   & 39.70 / 28.30 & 27.40 / 17.80 \\
 & Binary Search  & 51.70 / 36.20 & 6.90 / 6.90   & 44.10 / 30.10 & 24.00 / 16.60 \\
 & ECHO           & 68.40 / 67.90 & 28.10 / 26.80 & 68.80 / 67.20 & 28.80 / 27.20 \\
\midrule

Spectrum-based 
 & FAMAS          & 62.07 / \NA   & \textbf{41.38} / \NA & 55.56 / \NA   & 23.81 / \NA   \\
\midrule

\multirow{2}{*}{\shortstack[l]{Fine-tuning\\-based}} 
 & AgenTracer     & 69.10 / 63.82 & 20.70 / 20.68 & 69.62 / 63.73 & 42.90 / 37.30 \\
 & GraphTracer    & 74.91 / 69.74 & 28.63 / 27.97 & 76.64 / 67.42 & 49.97 / 44.35 \\
\midrule

\rowcolor{gray!10} 
\multicolumn{2}{l}{\textbf{\tool} (Ours)}  
 & \textbf{77.59} / \textbf{72.41} & 29.31 / \textbf{29.31} & \textbf{76.80} / \textbf{68.80} & \textbf{52.00} / \textbf{45.60} \\ 
\bottomrule
\end{tabular}
\vspace{-10pt}
\end{table*}

\subsection{Evaluation Metrics}
Following standard protocols~\cite{conf/icml/ZhangY0LHZL0W0W25, journals/corr/abs-2509-13782, journals/corr/abs-2509-03312}, we report accuracy at two granularities:
(1) \textit{Agent-level Accuracy}: The proportion of cases where the failure-responsible agent is correctly identified.
(2) \textit{Step-level Accuracy}: The proportion of cases where the exact root cause step is correctly attributed.
To mitigate randomness, all results represent the average of three independent runs using a strict top-1 criterion (i.e., the ground-truth target is ranked first).

\subsection{Implementation Details}
We implement {\tool} using Python 3.11. 
Unless otherwise specified, the base LLM employed is DeepSeek-V3.2 (thinking). 
For baselines, we prioritize official implementations with default configurations, while adopting reported results for those where reproduction was hindered by version discrepancies or randomness.
Experiments were conducted on a server with an Intel i7-10700 CPU, an NVIDIA TITAN RTX GPU and 32GB RAM.

\section{Results}
\label{sec:results}

\subsection{Main Results: Baseline Comparison}
\label{sec:baseline_comparison}
 
As shown in Table~\ref{tab:exp-results}, {\tool} demonstrates dominant performance over 8 baselines on \textit{Who\&When} benchmark, surpassing existing approaches in all metrics with only one minor exception.
{\tool} significantly surpasses all direct prompting baselines (\textit{All-at-once}, \textit{Step-by-step}, \textit{Binary Search}). 
While \textit{ECHO} benefits from hierarchical structure, it confines the usage of hierarchy to static context representation. 
{\tool} advances this paradigm by explicitly constructing a causal graph to guide top-down backtracking, enabling superior attribution accuracy.
Additionally, {\tool} outperforms expensive baselines with low computational costs.
While the spectrum-based \textit{FAMAS} relies on costly replays to achieve the best step-level accuracy on the hand-crafted subset (where longer trajectories enable reliable statistical analysis),  {\tool}'s dominance across all other settings proves that causal reasoning is more efficient, delivering robust performance with zero replay cost.
Similarly, fine-tuning methods (\textit{AgenTracer}, \textit{GraphTracer}) incur high training costs and struggle with generalization, yet still fall short of {\tool}.
This suggests that structuring flat logs into causal graphs unlocks LLM reasoning capabilities more effectively than parameter updates.

{\tool} is highly robust to the lack of ground truth ($\mathcal{G}$). 
While access to $\mathcal{G}$ benefits all methods, {\tool} consistently achieves superior results under identical settings, excluding \textit{FAMAS}'s step-level accuracy\footnote{As \textit{FAMAS} only reports results with ground truth in their paper, missing entries are denoted by {\NA} in Table~\ref{tab:exp-results}.}.
This stability is driven by our virtual oracle, which provides effective intermediate supervision, ensuring precise attribution without heavy reliance on the final task outcome.

\subsection{Cost Efficiency Analysis}
\label{sec:cost_analysis}


\begin{table}[t]
\centering
\small
\renewcommand{\arraystretch}{1}
\setlength{\tabcolsep}{9pt}

\caption{Average token cost per case on \textit{Who\&When} across two subsets in \textbf{w/ $\mathcal{G}$} setting. Lower is better.}

\label{tab:exp-costs}

\begin{tabular}{lcc}
\toprule
\multirow{2}{*}{\textbf{Method}} 
& \multicolumn{2}{c}{\textbf{Token Cost Per Case} $\downarrow$} \\
\cmidrule(lr){2-3}
& \textbf{Hand-Crafted} & \textbf{Alg.-Generated} \\
\midrule

All-at-Once    & 21,581  & 5,833 \\
Step-by-Step   & 87,720  & 6,533 \\
Binary Search  & 34,659  & 5,226\\
FAMAS          & 431,620 & 116,660\\
ECHO           & 53,701  & 25,642 \\
\rowcolor{gray!10} \textbf{\tool} & 55,085 & 19,504 \\
\bottomrule
\end{tabular}
\vspace{-15pt}
\end{table}

\begin{table*}[t]
\centering
\small
\renewcommand{\arraystretch}{1}
\setlength{\tabcolsep}{6pt}

\caption{Performance of {\tool} on the \textit{Who\&When} benchmark (\textbf{w/ $\mathcal{G}$}) using various base LLMs. We report agent- and step-level accuracy (\%) across both subsets, noting the model's thinking level in parentheses.}
\label{tab:exp-model}

\begin{tabular}{llcccc}
\toprule
\multirow{2}{*}{\textbf{Type}} & \multirow{2}{*}{\textbf{Base LLM}} 
& \multicolumn{2}{c}{\textbf{Hand-Crafted Dataset}} 
& \multicolumn{2}{c}{\textbf{Algorithm-Generated Dataset}} \\
\cmidrule(lr){3-4} \cmidrule(lr){5-6}
& & \textit{Agent-level} $\uparrow$ & \textit{Step-level} $\uparrow$
  & \textit{Agent-level} $\uparrow$ & \textit{Step-level} $\uparrow$ \\
\midrule

\multirow{3}{*}{Open Source}
& Qwen3-235B-A22B-Instruct-2507   & 63.79 & 22.41 & 69.04 & 32.53 \\
& Kimi-k2-0905-preview           & 67.24 & 24.13 & 68.25 & 42.06 \\
& Deepseek-V3.2(thinking)          & 77.58 & 29.31 & 76.80 & 52.00 \\
\midrule

\multirow{3}{*}{Closed Source}
& GPT-5.2(medium)       & 68.96 & 24.13 & 66.67 & 43.65 \\
& Claude-4.5-Sonnet(standard thinking)      & 68.96 & 27.58 & 63.49 & 51.58 \\
& Gemini-3-Flash-Preview(medium) & 70.68 & 29.31 & 69.84 & 50.79 \\
\bottomrule
\end{tabular}
\vspace{-15pt}
\end{table*}

To assess cost efficiency beyond accuracy, we present the average token consumption per failure log across different paradigms, shown in Table~\ref{tab:exp-costs}.
{\tool} incurs a moderate token increase ($2.5 \sim 3\times$) compared to direct prompting (e.g., \textit{All-at-once}), maintaining comparable costs to \textit{ECHO}. 
However, this cost is justified as {\tool} achieves significantly superior attribution accuracy.
Spectrum-based \textit{FAMAS} incurs high costs due to repeated replays, consuming \textbf{$6 \sim 8 \times$} more tokens than {\tool}. 
In contrast, {\tool} employs ``one-pass'' causal graph reasoning, bypassing costly environment re-interactions and drastically reducing overhead.
While fine-tuning methods (\textit{AgenTracer}, \textit{GraphTracer}) might offer lower inference costs\footnote{We do not report their inference costs as the fine-tuned models are not open-sourced.}, they demand high upfront costs for error-injection data generation and model fine-tuning. 
Conversely, {\tool} delivers superior performance directly off-the-shelf, bypassing the substantial overhead of the fine-tuning pipeline.

\subsection{Impact of Base LLM}
\label{sec:base_llm}

We evaluate {\tool}'s generalizability across diverse base LLMs, spanning closed-source models (e.g., GPT-5.2, Claude 4.5 Sonnet, Gemini-3) and open-weight models (e.g., Qwen 3, Kimi-k2, DeepSeek-V3.2), with results detailed in Table~\ref{tab:exp-model}.

Attribution accuracy positively correlates with the base LLMs' instruction-following and reasoning capabilities. 
The open-weight DeepSeek-V3.2, adopted as our default backbone, achieves the highest performance across all settings.
Among closed-source models, Gemini-3 leads on both subsets, except for Claude 4.5, which delivers the best step accuracy on the algorithm-generated subset.
Notably, closed-source models like GPT-5.2 underperform against high-thinking open models DeepSeek-V3.2. 
This is mainly because we configured them with lower thinking level to save costs. 
Furthermore, our prompt design might not be optimally adapted to specific models.
Collectively, these results demonstrate that {\tool} harnesses the base LLM's capabilities to construct precise causal graphs and virtual oracles, which in turn unlock the LLM's intrinsic reasoning potential for effective attribution.

\subsection{Ablation Study}
\label{sec:ablation}

To assess the individual impact of our proposed components, we define three modules: M1 (\textit{Hierarchical Causal Graph}), M2 (\textit{Hierarchical Oracle-Guided Backtracking}), and M3 (\textit{Counterfactual Attribution}). 
Building upon M1 as the structural basis (except the naive baseline, \textit{All-at-Once}), we evaluate three configurations:
\ding{182} Only M1: Utilizes the causal graph but performs attribution via direct prompting on graph without M2 and M3.
\ding{183} M1+M2: Incorporates virtual oracles for precise top-down backtracking but excludes M3.
\ding{184} M1+M3: Applies counterfactual reasoning directly on the graph but lacks the intermediate supervision and search space pruning provided by M2.


\begin{table}[t]
\centering
\small
\renewcommand{\arraystretch}{1}
\setlength{\tabcolsep}{8pt}

\caption{Ablation study of {\tool} on \textit{Who\&When} benchmark in \textbf{w/ $\mathcal{G}$} setting. We evaluate the incremental contribution of modules (M1--M3), reporting agent- and step-level accuracy (\%) across both subsets.}

\label{tab:exp-ablation}

\begin{tabular}{lcccc}
\toprule
\multirow{2}{*}{\shortstack[l]{\textbf{{\tool}}\\\textbf{Variants}}} 
& \multicolumn{2}{c}{\textbf{Hand-Crafted}} 
& \multicolumn{2}{c}{\textbf{Alg.-Generated}} \\
\cmidrule(lr){2-3} \cmidrule(lr){4-5}
& \textit{Agent} $\uparrow$ & \textit{Step} $\uparrow$ 
& \textit{Agent} $\uparrow$ & \textit{Step} $\uparrow$ \\
\midrule

All-at-Once     & 50.00 & 5.17  & 61.11 & 13.49 \\
Only M1         & 37.93 & 18.96 & 66.66 & 24.60 \\
M1+M2           & 51.72 & 22.41 & 61.11 & 34.12 \\
M1+M3           & 50.00 & 22.41 & 65.07 & 26.98 \\
\rowcolor{gray!10} \textbf{{\tool}} & \textbf{77.59} & \textbf{29.31} & \textbf{76.80} & \textbf{52.00} \\
\bottomrule
\end{tabular}
\vspace{-15pt}
\end{table}

The results are presented in Table~\ref{tab:exp-ablation}.
\textit{First}, introducing HCG (M1) alone consistently improves performance on the algorithm-generated subset, but yields mixed results on hand-crafted subset (characterized by longer trajectories): while step-level accuracy rises, agent-level accuracy declines compared to \textit{All-at-Once}.
This suggests that without guided reasoning (M2/M3), the dense structural information induces cognitive overload, hindering the LLM from effectively exploiting the graph.
\textit{Second}, incorporating M1 with either M2 or M3 recovers performance. 
M1+M2 enables precise backtracking but risks confusing symptoms with root causes due to the lack of causal attribution. 
Conversely, M1+M3 applies counterfactual reasoning but suffers from the absence of oracle-guided pruning, only relying on raw intuition to scan the entire graph. 
\textit{Third}, the full version of {\tool} achieves superior accuracy across all metrics, demonstrating the complementary role of each module.
M1 provides the structural foundation, M2 efficiently narrows the search space via top-down backtracking, and M3 rigorously verifies the root cause via counterfactual analysis.
This holistic integration is indispensable for precise attribution within lengthy and unstructured MAS trajectories.

\section{Conclusion}
\label{sec:conclusioin}

This paper presents {\tool}, a \textbf{C}ausal \textbf{HIE}rarchical \textbf{F}ailure attribution framework for LLM-based MASs.
By reconstructing flat logs into a hierarchical causal graph, {\tool} enables efficient top-down attribution via virtual oracle–guided backtracking and counterfactual reasoning.
Extensive experiments on \textit{Who\&When} benchmark demonstrate that {\tool} outperforms eight baselines on both agent- and step-level accuracy.
These results highlight the importance of causal structure for effective failure attribution in lengthy and unstructured MAS logs.

\section*{Limitations}

A primary limitation is that the effectiveness of {\tool} relies on the fidelity of the hierarchical causal graph and virtual oracles. Upstream inaccuracies (e.g., hallucinated edges) may propagate to the final diagnosis. Additionally, our evaluation is currently limited to the \textit{Who\&When} benchmark, the sole public dataset available, necessitating future validation on broader systems.
However, the performance advantage over eight representative baselines on this benchmark substantially alleviates these threats.
Second, {\tool} focuses on identifying single decisive root cause, aligning with the assumption in \textit{Who\&When} benchmark.
Whether our method works for cumulative error propagation, where failure results from a sequence of minor deviations rather than a single catastrophic error, remains to be verified in future studies.

\bibliography{custom}

\appendix
\section{Details for RAG-based Task Decomposition}
\label{appx:task_decomposition}
We detail the RAG component for task decomposition. We retrieve semantically similar exemplars from a reference knowledge base and inject them into the prompt as decomposition prototypes to encourage verifiable subtask stages.

The knowledge base is built from two public datasets: (1) GAIA, from which we utilize all 165 available instances that provides explicit step annotations (\texttt{Steps}) used as decomposition templates; (2) AssistantBench, from which we select 33 high-quality instances whose \texttt{explanation} fields contain rich implicit sub-goals and verification trails to serve as decomposition guidelines. During retrieval, we employ a fixed number of 2 exemplars for both the initial retrieval stage and the final prompt insertion. We compute cosine similarity over embeddings to identify relevant entries, while excluding any knowledge base instances originating from the current evaluation task to prevent data contamination.

\noindent\textbf{Knowledge Base Construction.}
To construct a unified knowledge base, we normalize entries from the GAIA and AssistantBench datasets into a retrievable plain-text format. Specifically, each GAIA entry is formed by concatenating the question (Question) with its reasoning steps (Steps), while each AssistantBench entry combines the task description (Task) with its detailed explanation (Explanation). This consistent text format facilitates efficient retrieval and utilization, as illustrated in \Cref{fig:kb_construction}.

\begin{figure*}
\begin{minipage}{0.98\textwidth}
\centering
\begin{lstlisting}[style=promptstyle]
    - GAIA entry format: {Question} + {Steps}
    - AssistantBench entry format: {Task} + {Explanation}
\end{lstlisting}
\captionof{figure}{Knowledge Base construction description and formatting.}
\label{fig:kb_construction}
\end{minipage}
\vspace{-5pt}
\end{figure*}

\noindent\textbf{Retrieved Example.} 
These retrieved exemplars illustrate different stylistic conventions in presenting tasks and their reasoning processes across benchmarks. The exemplar from AssistantBench uses a "Task" field for the main query followed by an "Explanation" field, where the reasoning implicitly incorporates sub-task decomposition and tool usage suggestions. In contrast, the GAIA exemplar employs a "Question" field for the task description and explicitly lists numbered "Steps" to delineate the sub-tasks in a structured, sequential manner. Such variations in format help the model adapt to diverse ways of structuring prompts and reasoning traces.Retrieved exemplars are shown in \Cref{fig:retrieved_examples}.

\begin{figure*}
\begin{minipage}{0.98\textwidth}
\centering
\begin{lstlisting}[style=promptstyle]
 [Injected exemplar 1]
 Source: AssistantBench
 Task: Which gyms (not including gymnastics centers) in West Virginia are within 5 miles (by car) of the Mothman Museum?
 Explanation: You can use Google Maps to find the Mothman Museum and then search nearby gyms within 5 miles, ...
    
 [Injected exemplar 2]
 Source: GAIA
 Question: A 5-man group made up of one tank, one healer, and three DPS is doing a dungeon ...
 Steps: 1. Searched "WoW classes" on Google. 2. Opened ... 3. Identified the relevant classes ... 4. Listed the classes in alphabetical order ...
\end{lstlisting}
\captionof{figure}{Retrieved exemplars injected into the prompt}
\label{fig:retrieved_examples}
\end{minipage}
\vspace{-8pt}
\end{figure*}

\noindent\textbf{Full Prompt.}
The prompt is designed to guide the LLM in performing RAG-based task decomposition by incorporating the original question, ground truth, multi-agent conversation history, and retrieved reference exemplars, thereby ensuring that the generated subtasks are both semantically meaningful and faithful to the actual execution trajectory. It further enforces a structured self-reflection process—consisting of draft optimization, evidence alignment, and final optimization.The full prompt template is provided in \Cref{fig:full_decomp_prompt}.

\begin{figure*}
\begin{minipage}{0.98\textwidth}
\centering
\begin{lstlisting}[style=promptstyle]
 You are an AI assistant tasked with analyzing a multi-agent conversation history when solving a real-world problem.
 The problem is: {question}
 The correct answer for the problem is: {ground_truth}
 Here is the conversation in JSON format: {history_text}
 There are total {len(history_text)} steps, each entry provides the agent output and its role.
 Here is the retrieved reference example: {rag_text}
 Based on this conversation and retrieved example, please decompose the reasoning into semantic subtasks.
 You must perform a self-reflection process to optimize your decomposition:
 1. Draft Optimization: propose an initial set of subtasks.
 2. Evidence Alignment: ensure each subtask's step range aligns with the dialogue.
 3. Final Optimization: ensure step ranges are continuous, cover all steps, and do NOT overlap.
 
\end{lstlisting}
\captionof{figure}{Prompt for RAG-based task decomposition.}
\label{fig:full_decomp_prompt}
\end{minipage}
\vspace{-10pt}
\end{figure*}


\begin{figure*}[t] 
    \centering
    \begin{subfigure}[b]{0.24\linewidth}
        \centering
        \includegraphics[width=\linewidth]{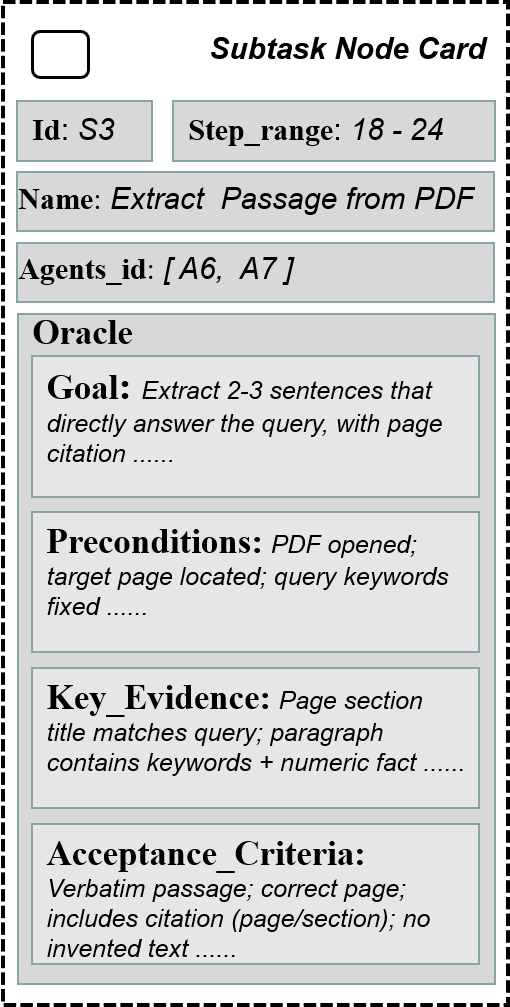} 
        \caption{Subtask node.}
        \label{subfig:graph_subtask}
    \end{subfigure}
    \hfill 
    \begin{subfigure}[b]{0.24\linewidth}
        \centering
        \includegraphics[width=\linewidth]{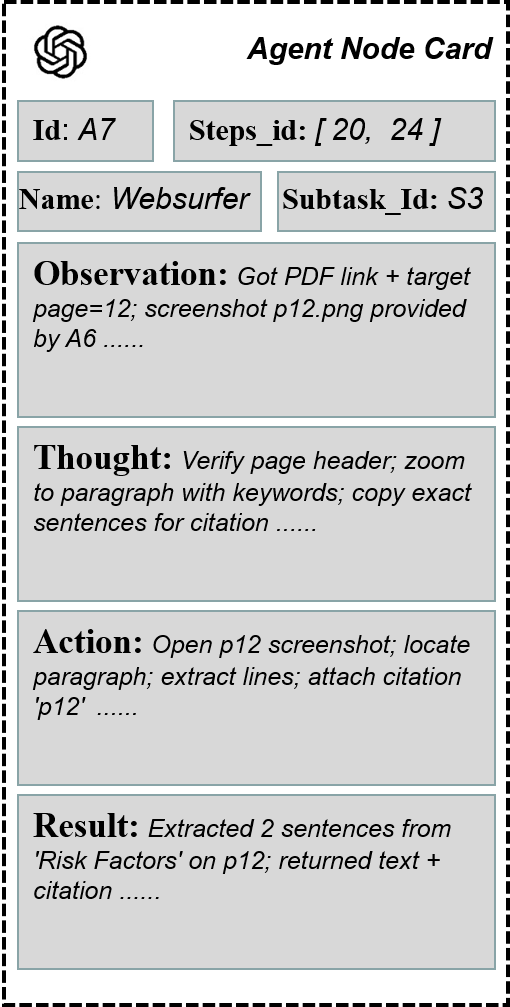} 
        \caption{Agent node.}
        \label{subfig:graph_agent}
    \end{subfigure}
    \hfill 
    \begin{subfigure}[b]{0.483\linewidth}
        \centering
        \includegraphics[width=\linewidth]{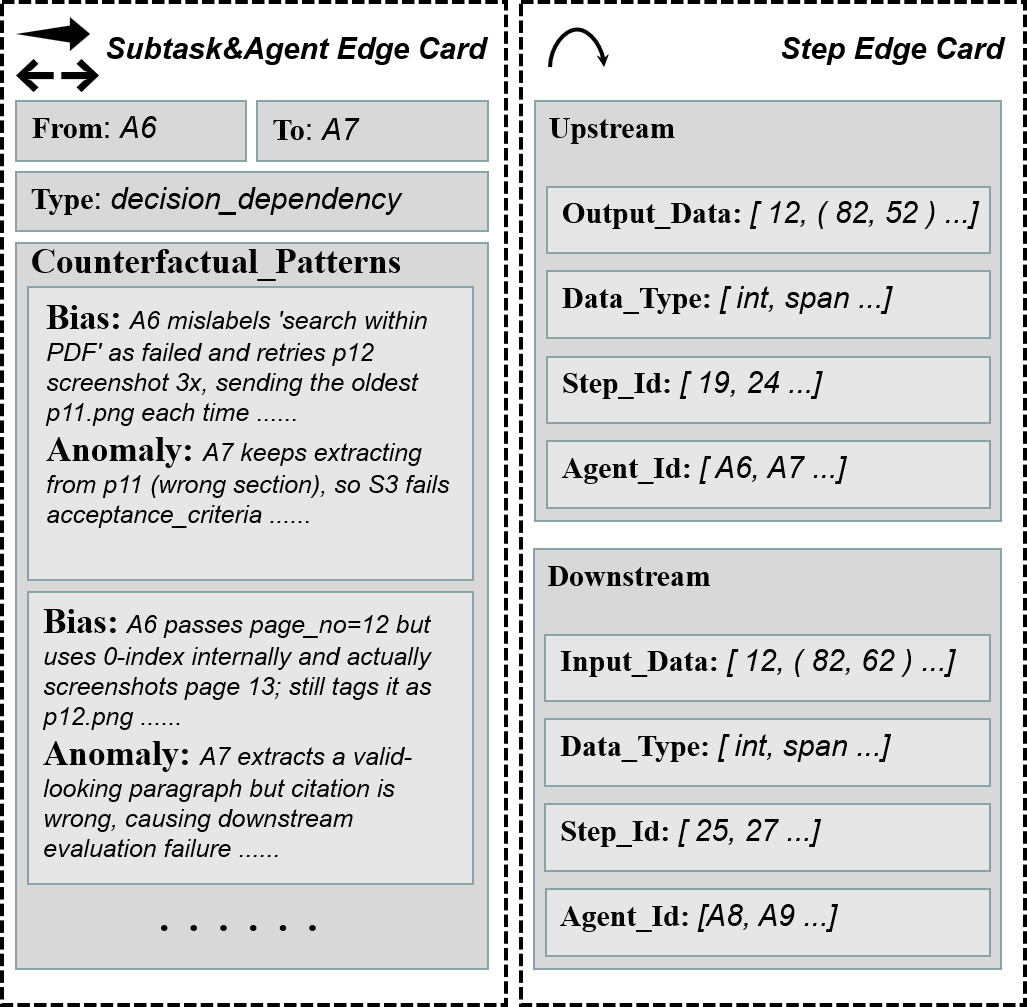} 
        \caption{Edges.}
        \label{subfig:graph_edge}
    \end{subfigure}
    
    \caption{An example of hierarchical causal graph.
    }
    \label{fig:graph}
    \vspace{-10pt}
\end{figure*}

\section{Prompt for OTAR Parsing}
\label{appx:prompt_otar_parsing}
The prompt is designed to guide the LLM in parsing multi-agent execution traces into structured OTAR (Observation, Thought, Action, Result) tuples for each agent within predefined subtasks, thereby enabling a rigorous characterization of agent behaviors that extends the original TAR schema. It enforces a strict output format that requires exact matching of subtask names, identification of active agents in each subtask’s step range, and detailed breakdown of their Observation, Thought, Action, and Result components directly extracted from the conversation history. The prompt itself employs a highly constrained, template-driven structure with clear sectional directives to ensure consistent and parseable OTAR annotations across all subtasks.The prompt template for OTAR parsing is provided in \Cref{fig:otar_prompt}.

\begin{figure*}
\begin{minipage}{0.98\textwidth}
\centering
\begin{lstlisting}[style=promptstyle]
 You are an AI assistant tasked with analyzing multi-agent execution traces.
 The problem is: {question}
 The correct answer for the problem is: {ground_truth}
 Here is the conversation in JSON format: {history_text}
 There are total {len(history_text)} steps, each entry provides the output of the agent and its role.
 Below are the subtasks: {subtasks_text}
 Your job:
 1. For each subtask, identify the agents that actively perform actions within its step_range.
 2. Summarize each agent's behavior into Action / Observation / Thought / Result.
 For EACH subtask, you MUST answer in the following strict format:
 The Subtask Name: (must exactly match one of the given subtask names)
 Agents:
 - Agent: <agent_name>
 -- Action: <what this agent did>
 -- Observation: <what they saw>
 -- Thought: <their reasoning>
 -- Result: <their output>
 (repeat '- Agent' blocks if multiple agents in this subtask)
\end{lstlisting}
\captionof{figure}{Prompt for OTAR parsing.}
\label{fig:otar_prompt}
\end{minipage}
\end{figure*}

\section{Prompt for Edge Construction}
\label{appx:edge_construction}
\noindent\textbf{Subtask \& Agent Edge Prompt.} To construct the structured dependencies for the multi-level causal graph, this prompt is specifically designed to generate Subtask Edges and Agent Edges. Its core mechanism lies in defining specific edge types and Counterfactual Patterns, explicitly delineating the logical progression between subtasks and the collaboration and error propagation pathways among agents within a subtask. The prompt template for subtask \& agent edge construction is provided in \Cref{fig:subtask_agent_edge_prompt}.

\begin{figure*}
\begin{minipage}{0.98\textwidth}
\centering
\begin{lstlisting}[style=promptstyle]
 You are an expert in causal reasoning and multi-agent task analysis.
 The problem is: {question}
 The correct answer for the problem is: {ground_truth}
 Here is the conversation in JSON format: {history_text}
 There are total {len(history_text)} steps, each entry provides the output of the agent and its role.
 Below are the subtasks with their agents: {subtasks_agents_text}
 Your job:
 1. Construct causal edges ONLY for consecutive subtask pairs: (S1->S2), (S2->S3), ... (subtask-subtask edges)
 2. For each subtask, construct causal edges BETWEEN agents inside this subtask only (no cross-subtask agent edges)
 3. Use the agent-level DAG to describe how observations, reasoning, and decisions flow from one agent to another
 For EACH subtask-subtask edge (consecutive pairs) and EACH agent-agent edge (per subtask), 
 you MUST output ONE block in the following exact format:
 From: <upstream subtask id (e.g., S1) or upstream agent name>
 To: <downstream subtask id (e.g., S2) or downstream agent name>
 Type: <subtask edge type: data_dependency / logical_prereq; or agent edge type: obs_dependency / reasoning_continuation / decision_dependency / environment_feedback / memory_ref / loop_control>
 Counterfactual_Patterns:
 - Bias: ...
   Anomaly: ...
   
 Guidelines:
 - You may output zero, one, or multiple Failure Modes items for each edge
 - Output all subtask edges first, then agent edges by subtask
 Additional Guidelines for Counterfactual Pattern Construction:
 1. Counterfactual patterns must correspond to 4 attribution scenarios: Local (whether the error of the current edge is caused by itself), Planning-Control (planner/executor responsibility in control flow/loops), Data-Flow (the starting point of data corruption in data streams), and Deviation-Aware (whether the deviation is reversible).
 2. For Subtask/Agent edges, explicitly bind the potential bias of the upstream node (Bias) to the observable anomaly of the downstream node (Anomaly), e.g., "If the Thought of the upstream Agent has hallucinations (Bias), the Result of the downstream Agent will have data errors (Anomaly)".
 3. When constructing Planning-Control counterfactuals, distinguish between planner responsibility (repeating failed strategies) and executor responsibility (execution anomalies under valid strategies).
 4. When constructing Data-Flow counterfactuals, associate the consistency corruption node of specific data items.
 5. When constructing Deviation-Aware counterfactuals, mark whether the deviation is reversible (whether subsequent steps self-correct).

\end{lstlisting}
\captionof{figure}{Prompt for subtask \& agent edge construction.}
\label{fig:subtask_agent_edge_prompt}
\end{minipage}
\end{figure*}

\noindent\textbf{Step Edge Prompt.}To complement the abstract causal patterns with concrete data evidence, this prompt is specifically designed to construct Step Edges. Its core mechanism lies in strictly matching the output data from upstream steps with the input data of downstream steps, explicitly capturing the concrete data flow between execution steps to provide definitive evidence trails for causal backtracking. The prompt template for step edge construction is provided in \Cref{fig:step_edge_prompt}.

\begin{figure*}
\begin{minipage}{0.98\textwidth}
\centering
\begin{lstlisting}[style=promptstyle]
 You are an AI assistant tasked with analyzing multi-agent execution traces.
 The problem is: {question}
 The correct answer for the problem is: {ground_truth}
 Here is the conversation in JSON format: {history_text}
 There are total {len(history_text)} steps, each entry provides the output of the agent and its role.
 Below are the subtasks: {subtasks_text}
 Your job:
 1. Identify step-level edges (meaningful data passing between steps)
 2. For each step edge, identify the upstream step (data producer) and downstream step (data consumer)
 You MUST answer in the following strict format:
 - Upstream: <integer step id where the data is produced>
   output_data: "short description of the data (e.g., 'distance=30' or 'parsed_goal')"
   data_type: "text/numeric/list/boolean"
   step_id: <step id in the upstream step>
   agent_id: <agent id in the upstream step>
 - Downstream: <integer step id where the data is used>
   input_data: "short description of the data (e.g., 'distance=30' or 'parsed_goal')"
   data_type: "text/numeric/list/boolean"
   step_id: <step id in the downstream step>
   agent_id: <agent id in the downstream step>
    
 Guidelines:
 - Step Edges should capture meaningful data passing from upstream to downstream step
 - You may output zero, one, or multiple Step Edges items
    
\end{lstlisting}
\captionof{figure}{Prompt for step edge construction.}
\label{fig:step_edge_prompt}
\end{minipage}
\end{figure*}

\section{Prompt for Oracle Synthesis}
\label{appx:oracle_synthesis}
To generate the intermediate supervision signals for hierarchical backtracking, this prompt is specifically designed to synthesize Subtask Virtual Oracles. Its core mechanism is to leverage the constraints from previously generated oracles and the context of subsequent unprocessed trajectories, employing sequential generation and global self-checking to construct structured and internally consistent verification criteria for each subtask. The prompt template for oracle synthesis is provided in \Cref{fig:oracle_synthesis_prompt}.

\begin{figure*}
\begin{minipage}{0.98\textwidth}
\centering
\begin{lstlisting}[style=promptstyle]
 You are an AI assistant tasked with synthesizing a complete set of Subtask Virtual Oracles for hierarchical failure diagnosis in a multi-agent trajectory.
 The problem is: {question}
 The correct answer for the problem is: {ground_truth}
 Here is the conversation in JSON format: {history_text}
 There are total {len(history_text)} steps, each entry provides the output of the agent and its role.
 Here is the retrieved reference example: {rag_text}
 Given subtask plan produced by the decomposition stage: {subtasks}
 Your goal: Generate ALL virtual oracles for ALL subtasks.
 IMPORTANT thinking rule:
 - You must synthesize oracles in subtask order: k = 1..K.
 - While generating the oracle for the current subtask, treat all previously generated oracles as "previous oracle constraints" (internal memory). Use them to ensure consistency, avoid contradictions, and define valid preconditions.
 You MUST perform a self-check process:
 1) Draft All Oracles (sequential):
    - For k=1..K, draft the oracle using the Problem, the retrieved reference example, the subsequent unprocessed trajectory, and the previously generated oracle constraints.
 2) Global Consistency Check:
    - No oracle contradicts the problem instruction or earlier oracles.
    - Precondition must only depend on information that is available before or at the beginning of this subtask (i.e., upstream outputs, environment states, or outcomes implied by earlier oracles).
    - Key Evidence must include only essential evidence that should be checked during this subtask.
    - Acceptance Criteria must be checkable after execution and falsifiable.
 3) Finalize All Oracles.

 Output format:
 For each subtask, output exactly the following block, repeated K times in order:
 -Subtask Name: <copy exactly from subtask plan>
 -Oracle:
  Goal: <what this subtask should achieve>
  Precondition: <each item must reference upstream outputs/environment states only>
  Key Evidence: <critical facts/claims/tool-return fields/intermediate  quantities to verify>
  Acceptance Criteria: <checkable post-hoc; define pass/fail>
\end{lstlisting}
\captionof{figure}{Prompt for Oracle Synthesis.}
\label{fig:oracle_synthesis_prompt}
\end{minipage}
\end{figure*}

\section{Prompt for Hierarchical Backtracking}
\label{appx:hierarchical_backtracking}
To achieve fault localization from coarse to fine granularity, this prompt is specifically designed to perform Hierarchical Backtracking. Its core mechanism is to sequentially conduct semantic evaluation and comparison at three levels (subtask, agent, and step) by leveraging the causal graph and virtual oracles, progressively narrowing down the candidate error nodes to pinpoint the root cause. The prompt template for hierarchical backtracking is provided in \Cref{fig:hierarchical_backtracking_prompt}.

\begin{figure*}
\begin{minipage}{0.98\textwidth}
\centering
\begin{lstlisting}[style=promptstyle]
 You are an AI assistant tasked with analyzing a multi-agent conversation solving a real-world problem and performing hierarchical failure backtracking.
 The problem is: {question}
 The correct answer for the problem is: {ground_truth}
 Here is the conversation in JSON format: {history_text}
 There are total {len(history_text)} steps, each entry provides the output of the agent and its role.
 Here is the causal graph describing the hierarchical reasoning structure:{graph}
 Your job :
 1) Subtask-level backtracking:
   - Traverse subtasks in reverse topological order according to the subtask edges in the graph.For each subtask, determine its actual execution output from the conversation slice within its step range.Compare the actual execution output against this subtask's oracle Goal and Acceptance Criteria.Internally make a binary discrepancy decision (0/1). If discrepancy=1, include this subtask in the candidate error subtasks set.
 2) Agent-level backtracking:
   - For each candidate subtask, evaluate each constituent agent.Compare the agent OTAR against the oracle Preconditions and Key Evidence of that subtask.Internally make a binary discrepancy decision (0/1). If discrepancy=1, include this agent in the candidate error agents set.
 3) Step-level backtracking:
   - For each candidate agent within a candidate subtask, evaluate its executed steps in that subtask.
   - For each step, extract concrete execution details (prioritize tool input/output, intermediate computed values, cited facts, or explicit variable references).
   - Cross-check the step's execution details against:
     (i) the agent's OTAR summary
     (ii) the full oracle checklist of this subtask (Goal, Preconditions, Key Evidence, Acceptance Criteria).
   - Internally make a binary discrepancy decision (0/1). If discrepancy=1, include this step as a candidate error step.
 Now, please strictly follow this output format:
 Candidate Error Subtasks: [Id1, Id2, ...]
 Candidate Error Agents: [agent1, agent2, ...]
 Candidate Error Steps: [step_Id1, step_Id2, ...]

\end{lstlisting}
\captionof{figure}{Prompt for hierarchical backtracking.}
\label{fig:hierarchical_backtracking_prompt}
\end{minipage}
\vspace{-30pt}
\end{figure*}

\section{Details for Counterfactual Attribution}
\label{appx:attribution}

This appendix provides two illustrative examples for \textit{Planning-Control Attribution}. Both examples exhibit cyclic failures, but the responsibility differs: one is attributed to the \textit{planner} (no meaningful strategy update), and the other to the \textit{executor} (valid strategy shifts but abnormal execution). 

\begin{figure*}[t] 
    \centering
    \begin{subfigure}[b]{0.495\linewidth}
        \centering
        \includegraphics[width=1.02\linewidth]{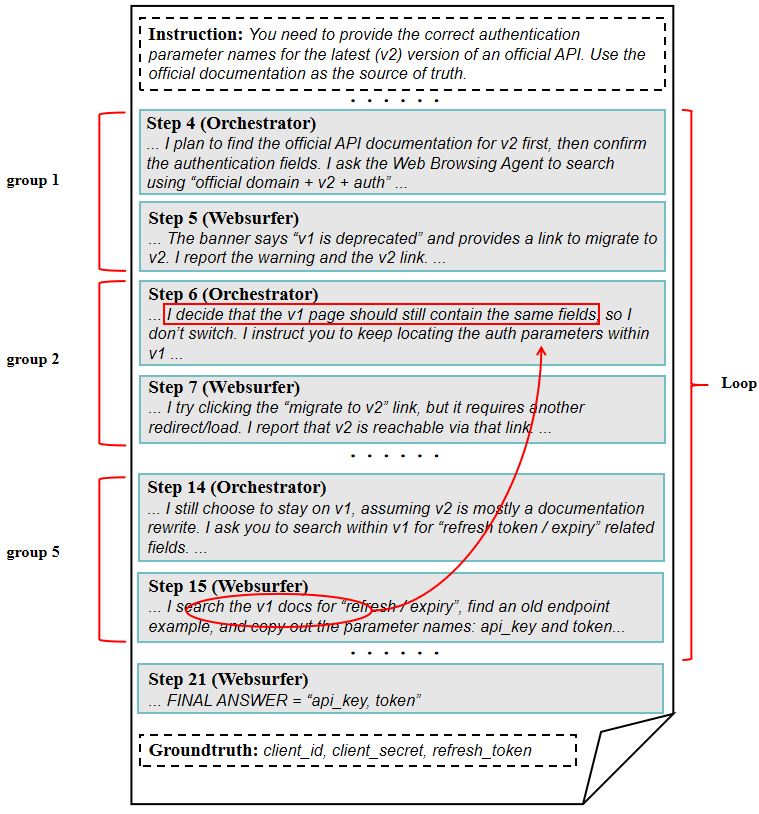} 
        \caption{An example of blaming the planner.}
        \label{subfig:loop_planner}
    \end{subfigure}
    \hfill 
    \begin{subfigure}[b]{0.495\linewidth}
        \centering
        \includegraphics[width=1.02\linewidth]{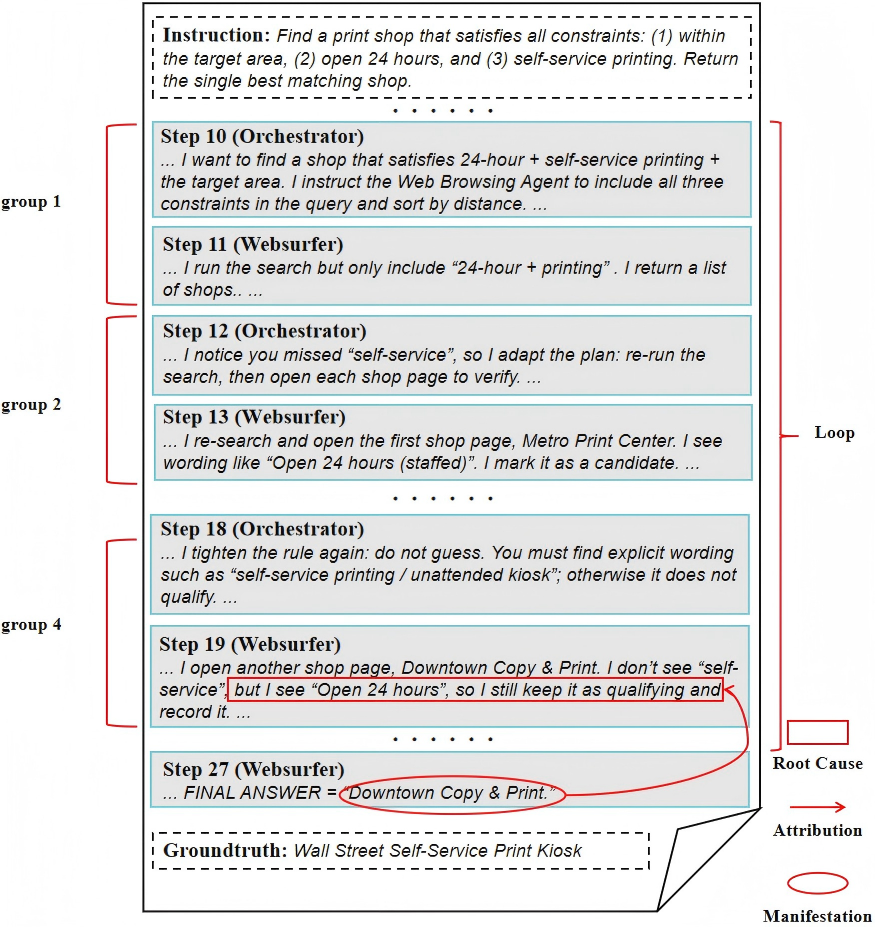} 
        \caption{An example of blaming the executor.}
        \label{subfig:loop_executor}
    \end{subfigure}
    \hfill 
    
    \caption{Examples of Planning-Control Attribution .}
    \label{fig:loop}
\end{figure*}

In Fig.~\ref{subfig:loop_planner}, at step 4 the planner issues a correct intent: to query the information about version v2. WebSurfer subsequently retrieves the critical evidence that v1 has been deprecated and provides a link to v2. Under a correct control flow, the planner should then instruct the executor to follow the v2 link. However, at step 6 the planner remains fixated on v1 and generates a deviated instruction to continue searching around v1, which drives the trajectory into cyclic behavior and yields multiple subsequent loop groups. Across these repetitions, despite receiving repeated failure signals, the planner continues to emit semantically equivalent “explore v1” thoughts or commands, indicating a lack of effective strategy update. Eventually, WebSurfer is forced to produce an incorrect answer grounded in v1. Therefore, this case satisfies our criterion of persisting with a failed approach under failure feedback, and we attribute the root cause to \textit{Planner Responsibility}.

In Fig.~\ref{subfig:loop_executor}, at step 10 the planner explicitly issues a query instruction and emphasizes the full set of constraints: “24-hour + self-service printing + the target area.” Yet at step 11, WebSurfer ignores the “self-service” constraint and returns an invalid candidate. In response, at step 12 the planner identifies the omission and reasserts the constraints; nevertheless, at step 13 WebSurfer repeats a similar execution deviation, pushing the trajectory into another loop and forming multiple loop groups. Unlike Figure 12a, the planner repeatedly detects the error and proposes reasonable strategy shifts intended to break the loop, while the executor persistently ignores or misinterprets parts of the constraints and thus keeps producing abnormal results. Hence, this case matches our criterion of valid planning but abnormal execution, and we attribute the root cause to \textit{Executor Responsibility}.

Moreover, the prompt template for counterfactual attribution systematically integrates four core attribution strategies: it first employs Local Attribution to determine if the error originates locally. If not, it then utilizes Planning-Control Attribution to dissect responsibility within cyclic behaviors, and Data-Flow Attribution to trace back to the source of data corruption. Finally, Deviation-Aware Attribution acts as a validity filter to dismiss transient deviations that are later self-corrected by the system. The full prompt template for counterfactual attribution is provided in \Cref{fig:counterfactual_attribution_prompt}.

\begin{figure*}
\begin{minipage}{0.98\textwidth}
\centering
\begin{lstlisting}[style=promptstyle]
 You are an AI assistant tasked with analyzing a multi-agent conversation solving a real-world problem and performing counterfactual failure attribution.

 The problem is: {question}
 The correct answer for the problem is: {ground_truth}
 Here is the multi-agent conversation: {history_text}
 There are total {len(history_text)} steps, each entry provides an agent's output.
 Here is the structured candidate_set: {candidate_set}

 Here is the graph: {dag_graph}

 Your job is to identify the SINGLE most responsible reasoning mistake that directly leads to the wrong final result.

 You MUST follow this internal reasoning procedure over the candidate_error_steps :

 Stage A Local Attribution (local vs upstream propagation):
 - For each candidate step x:
   - Use the graph's predecessor relations and edge-attached counterfactual patterns to judge whether there exists any upstream step that can causally explain the anomaly at x under the oracle constraints.
   - If no upstream causal trigger can explain x AND x received valid inputs yet produced an incorrect output, treat x as a strong local-origin root-cause candidate.

 Stage B Planning-Control Attribution (control / loop responsibility):
 - Use loop information from the graph (e.g., loop groups and entry/internal/exit roles) to analyze whether the failure is caused by redundant cyclic behavior.
 - Distinguish planner vs executor responsibility:
   - Planner responsibility: despite repeated error signals, the planner repeats semantically identical thoughts/commands and fails to adapt strategy.
   - Executor responsibility: the planner proposes valid strategy shifts, yet execution still yields abnormal results.

 Stage C Data-Flow Attribution (data dependency responsibility):
 - Use data-flow information in the graph to trace how key data items are produced and consumed.
 - Decide whether the candidate step:
   - fabricates data (no upstream basis),
   - misinterprets upstream data,
   - misuses otherwise correct data.
 - If the error propagates via data, prefer attributing responsibility to the earliest step where valid upstream inputs are first corrupted into an abnormal result .

 Stage D Final Screening (Reversibility and Irrecoverability):
 - Deviation-aware filter (reversibility): Check whether the deviation introduced by a candidate step is later self-corrected such that oracle constraints are re-satisfied (system returns to a valid state). If the deviation is reversible (self-corrected later), assign it minimal responsibility compared to irreversible deviations.
 - Irrecoverability tie-break (first point blocking recovery): Prefer the candidate step that first makes it hard or impossible to restore the correct reasoning path through conventional means, rather than the earliest deviation.

 Final decision rule;
 - Select ONE step that best satisfies: (1) true origin (local-origin or earliest data corruption) AND (2) passes Final Screening (irreversible and/or first irrecoverable point) AND (3) strongest downstream impact / irrecoverability.
 - After deciding the single most responsible step:
   - Determine the agent name at that step (from candidate_set and conversation context).
   - Determine the exact step index (step_id).
   - Provide a concise explanation (2-3 sentences) that explicitly mentions which attribution stages (Local / Planning-Control / Data-Flow / Final Screening) support this choice.

 Now, please answer in this exact plain text format:
 Agent Name: (your final prediction, a single agent name)
 Step Number: (your final prediction, a single integer step id)
 Reason for Mistake: (your explanation, summarize in 2-3 sentences)
 No special symbols, no extra commentary.

\end{lstlisting}
\captionof{figure}{Prompt for counterfactual attribution.}
\label{fig:counterfactual_attribution_prompt}
\end{minipage}
\end{figure*}

\end{document}